\documentclass[10pt,twocolumn,letterpaper]{article}

\usepackage{cvpr}              
\usepackage[accsupp]{axessibility} 

\usepackage[dvipsnames]{xcolor}

\usepackage{graphicx}

\usepackage[pagebackref=true,breaklinks=true,colorlinks,bookmarks=false]{hyperref}
\usepackage{amsmath}
\usepackage{amssymb}

\usepackage{multirow}
\usepackage{multicol}
\usepackage{diagbox}
\usepackage{xfrac}

\usepackage[labelformat=simple,subrefformat=simple]{subcaption}
\usepackage{caption}
\usepackage{algorithm}
\usepackage{algpseudocode}


\def\paperID{*****} 
\def\confName{CVPR}
\def\confYear{2025}

\title{
Separating Shared and Domain-Specific LoRAs
for Multi-Domain Learning
}

\author{Yusaku Takama,
Ning Ding, Tatsuya Yokota, Toru Tamaki\\
Nagoya Institute of Technology\\
Nagoya, Japan}
\begin{document}
\maketitle

\begin{abstract}
    
Existing architectures of multi-domain learning have two types of adapters: shared LoRA for all domains and domain-specific LoRA for each particular domain. However, it remains unclear whether this structure effectively captures domain-specific information. In this paper, we propose a method that ensures that shared and domain-specific LoRAs exist in different subspaces; specifically, the column and left null subspaces of the pre-trained weights. We apply the proposed method to action recognition with three datasets (UCF101, Kinetics400, and HMDB51) and demonstrate its effectiveness in some cases along with the analysis of the dimensions of LoRA weights.

\end{abstract}

\section{Introduction}

With the recent development of deep learning, many models have been proposed for various tasks.
However, if there is a domain shift among datasets \cite{Torralba_Unbiased_dataset_bias_CVPR2011}, 
performance will deteriorate significantly.
Therefore, 
methods have been proposed to simultaneously train a model on multiple datasets of multiple domains, 
such as 
multi-domain learning \cite{Li_Efficient_Multi_Domain_CVPR2019,Rebuffi_Efficient_Multi_Domain_CVPR2018},
multi-dataset learning \cite{Zhou_Simple_MultiDataset_CVPR2022,Liang_Multi_Dataset_NeurIPS2022},
multi-task learning \cite{Vandenhende_Dence_MTL_survey_PAMI2021,Crawshaw_MTL_survey_2020,Zhang_MTL_Survey_2021},
and multimodal learning \cite{Likhosherstov_Polyvit_2021,Girdhar_Omnivore_CVPR2022,Xu_Multimodal_survey_PAMI2023,Xie_Large_Multimodal_Agents_survey_2024}.
These methods have a structure that uses a single model to learn multiple domains, datasets, modalities, and tasks. So, they can learn diverse feature representations by incorporating information from multiple datasets, not just from a single dataset. In this paper, we will refer to these learning approaches collectively as \emph{multi-domain} learning.

The cost of training models, however, has increased due to the large scale of recent models, highlighting the growing importance of efficient learning, especially in multi-domain learning. Therefore, 
research into parameter-efficient fine-tuning (PEFT) \cite{Han_PEFT_survey_2024,Xin_PEFT_survey_2024,Xing_PEFT_survey_2024}
has gained popularity;
PEFT fine-tunes only a subset of parameters,
such as additional adapters \cite{Zhang_LLaMA_Adapter_ICLR2024,Xin_VMT_Adapter_AAAI2024,Sung_VL_Adapter_CVPR2022,rebuffi_arXiv2017_learningmultiplevisualdomains,Houlsby_PMLR2019_Parameter-Efficient} and LoRAs  \cite{hu_arXiv2021_lora_low_rank_adaptation_large,Dettmers_QLoRA_2023}
instead of all weights, thus reducing computational costs.

While PEFT has been explored for single-domain problems,
some work applied PEFT to multi-domain learning \cite{Agiza_CVPR2024_MTLoRA,Xin_VMT_Adapter_AAAI2024,Liu_Polyhistor_NeurIPS2022,Karimi_Hyperformer_ACL2021}.
This approach enables more efficient learning with fewer parameters compared to training a separate model for each domain. 
Existing methods generally consist of two types of modules: a shared module and a domain-specific module. 
The shared module is a LoRA or adapter that learns domain-agnostic information across different domains, while
the domain-specific module captures domain-specific information for a particular domain.

However, in the structure of existing PEFT-based multi-domain learning, it is not clear whether the shared module effectively extracts information common to all domains or whether the domain-specific module extracts information specific to each domain. This is because the only difference between the two types of modules is whether they are trained on data in all domains or only in specific domains, and they do not have a structure that explicitly separates information.
In other words, compared to PEFT structures for a single domain, the performance improvements by the PEFT structures for multi-domain learning might merely be due to the increased number of learnable parameters in the shared module. 
In an extreme case, the shared module may also learn domain-specific information, while the domain-specific module may not work at all. 
In the previous work \cite{Agiza_CVPR2024_MTLoRA,Xin_VMT_Adapter_AAAI2024}, this separation was not incorporated or analyzed. If domain-agnostic and domain-specific information can be clearly separated, it is expected to be beneficial not only for improving model performance but also for developing new methods for understanding the information generalized to various tasks and domains.

\begin{figure}[t]
    \centering
    \includegraphics[width=1\linewidth]{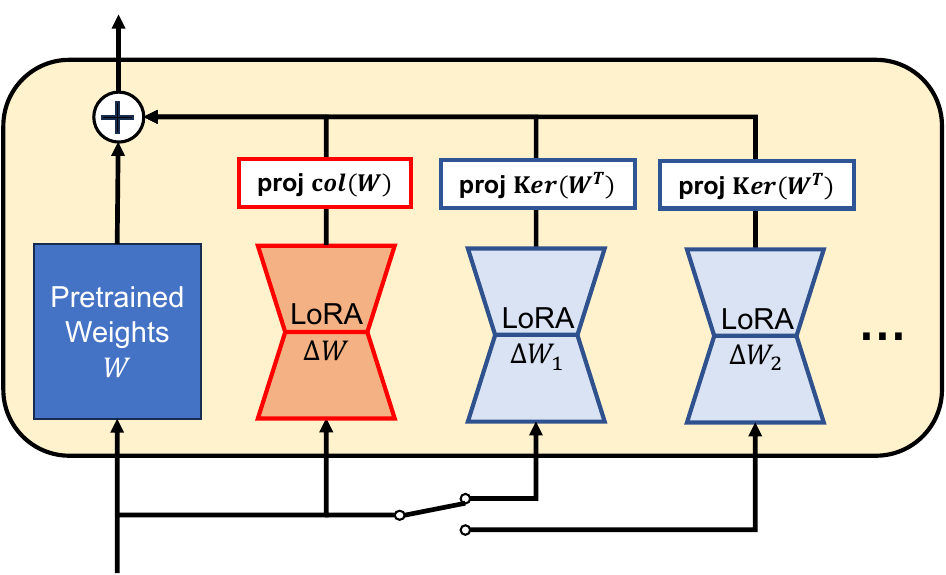}
    \caption{
    Overview of the proposed model structure with separated parameter spaces for learning the shared and domain-specific LoRA, where the domain-specific LoRA is learned independently for each domain.
    }
    \label{fig:inmodel} 
\end{figure}

To address this problem, we propose a method that explicitly separates the information learned by the shared and domain-specific modules by introducing constraints on the parameter space for each module.
The concept of the proposed method is shown in Figure~\ref{fig:inmodel},
where the pre-trained weight is $W$, the shared LoRA is $\Delta W$, and the domain-specific LoRA for domain $i$ is $\Delta W_i$.
We achieve separation by explicitly constraining the parameter space in which $\Delta W$ and $\Delta W_i$ are learned to the \emph{column space} and the \emph{left null space} of $W$, respectively. 
By constraining shared LoRA updates to the column space, which comprises most of the components of $W$ (the subspace with 95\% of principal components in this paper), we can enhance the knowledge contained in the pre-trained weights $W$ and effectively learn important features common across all domains. Furthermore, by constraining LoRA updates to the left null space, which contains very few components of $W$, we can efficiently learn new domain-specific information while avoiding interference with existing knowledge. 
In other words, by imposing constraints that ensure that shared and domain-specific LoRA are trained in mutually orthogonal spaces, we guarantee that shared and domain-specific LoRA are learned in different spaces. For different domains $i$ and $j$, we also impose constraints to ensure that the domain-specific modules for different domains are separated within the left null space.
This enables information from different domains to be learned independently.

In this work, we focus on multi-domain learning for action recognition with PEFT of a model pre-trained on an image dataset. 
Action recognition is the task of identifying human actions in videos and is expected to be utilized in various applications \cite{Selva_PAMI2023_Video_Trans_Survey, Kong_IJCV2022_Action_Recognition_Survey,Ulhaq_arXiv2022_Transformers_Action_Recognition_Survey},
and large-scale action recognition datasets \cite{kay_arXiv2017_kinetics400, Soomro_arXiv2012_UCF101, Kuehne_ICCV2011_HMDB51} are used for the supervised learning of these models. Recent approaches extend models pre-trained on large-scale image datasets, such as recent Vision Transformers (ViTs) \cite{Dosovitskiy_ICLR2021_ViT_Vision_transformer} and CLIP \cite{Radford_ICML2021_CLIP}. In this case, 
if a structure can separate information in the video domains
learned in $\Delta W_i$ from the pre-trained weights $W$ learned in the image domain, this could contribute to addressing the issue of static bias \cite{Weinzaepfel_IJCV2021_Mimetics_dataset,Li_REPAIR_CVPR2019,Ferrari_RESOUND_ECCV2018}, which has become a growing concern in recent years.

\section{Related work}

\subsection{PEFT}

Parameter-Efficient Fine-Tuning (PEFT) \cite{Han_PEFT_survey_2024,Xin_PEFT_survey_2024,Xing_PEFT_survey_2024} is a technique for efficiently fine-tuning pre-trained large-scale models. In conventional fine-tuning, which updates all the parameters of a large-scale model, issues such as computational cost and memory usage become significant challenges. In contrast, PEFT enables efficient and flexible fine-tuning by updating only a subset of the parameters.

One of the typical PEFT methods is adapters; a trainable module inserted into a pre-trained frozen model, which is expected to be small compared to the original model. Many existing methods are specialized for specific architectures, such as the residual adapter \cite{rebuffi_arXiv2017_learningmultiplevisualdomains} for ResNet \cite{He_CVPR2016_ResNet} or the adapter module \cite{Houlsby_PMLR2019_Parameter-Efficient} for BERT \cite{Devlin-ACL2019-BERT}.

The other type is the Low-Rank Adapter (LoRA) \cite{hu_arXiv2021_lora_low_rank_adaptation_large}, which is often used in image generation tasks \cite{Rombach_2022_CVPR_stable_diffusion}. It inserts a linear adapter consisting of two low-rank matrices in parallel with the linear layer of ViT, significantly reducing the number of trainable parameters.
Unlike general adapters, it has the advantage of almost no increase in computational cost during inference.

\subsection{Multi-domain learning}

There are many learning approaches that handls diverse datasets and tasks within a single model, such as multi-domain learning \cite{Li_Efficient_Multi_Domain_CVPR2019,Rebuffi_Efficient_Multi_Domain_CVPR2018}, multi-dataset learning \cite{Zhou_Simple_MultiDataset_CVPR2022,Liang_Multi_Dataset_NeurIPS2022}, multi-task learning \cite{Vandenhende_Dence_MTL_survey_PAMI2021,Crawshaw_MTL_survey_2020,Zhang_MTL_Survey_2021}, and multimodal learning \cite{Likhosherstov_Polyvit_2021,Girdhar_Omnivore_CVPR2022,Xu_Multimodal_survey_PAMI2023,Xie_Large_Multimodal_Agents_survey_2024}. 
In this paper, we refer to these learning approaches as multi-domain learning.

The aim of multi-domain learning is to achieve a better performance in each domain while also enhancing feature representations by incorporating diverse information across different domains. 
However, training a model for multiple datasets
becomes difficult in terms of computational cost and memory usage during training and inference. To address this issue, 
multi-domain learning utilizing PEFT has been proposed, as described below.

\subsection{Adapting PEFT for multi-domain learning}

Examples of multi-domain learning with PEFT include MTLoRA \cite{Agiza_CVPR2024_MTLoRA} and VMT-Adapter \cite{Xin_VMT_Adapter_AAAI2024}. These methods have a structure that contains
the shared adapters for all domains, as well as
the domain-specific adapters for each domain which
are trained only when a sample batch comes from the corresponding domain.

However, as mentioned before, the structure of these methods does not explicitly separate the information learned by the shared and domain-specific adapters. Therefore, it is not clear whether the shared adapters extract information common to all domains or whether the domain-specific adapters extract information specific to each domain. 
In this study, we propose a structure that explicitly separates these two types of adapters, by extending the structure of MTLoRA \cite{Agiza_CVPR2024_MTLoRA}.

\section{Proposed method}

The proposed method introduces constraints to separate pre-trained weights $W$, shared LoRA $\Delta W$, and domain-specific LoRAs $\Delta W_i$,
with a structure of \cite{Agiza_CVPR2024_MTLoRA}. 
That is, during training, we train the shared LoRA $\Delta W$ in parallel with $W$ for data of any domain, and train the domain-specific LoRA $\Delta W_i$ for data of the $i$-th domain only.
Here, LoRA weights are decomposed as the product of low-rank matrices $A, B$ with a predefined rank $r$; $\Delta W = B A^T$ and $\Delta W_i = B_i A_i^T$,
where $W, \Delta W, \Delta W_i \in \mathbb{R}^{d \times d'}$,
$A, A_i \in \mathbb{R}^{d' \times r}$ and $B, B_i \in \mathbb{R}^{d \times r}$.
Then, we used all the LoRAs combined as $(W + \Delta W + \Delta W_i) x$ for input $x \in \mathbb{R}^{d'}$ of domain $i$ during inference.

The constraints proposed in this paper utilize the left null space and column space of $W$, which will be explained in the next section. Then we introduce constrains the LoRA weights to the column and left null spaces.

\subsection{Preliminary}
\subsubsection{Column space and left null space}

The null space $\operatorname{Ker}(W)$ of a $d \times d'$ matrix $W$ is defined by
\begin{align} 
\mathrm{Ker}(W) = \{ \boldsymbol{x} \in \mathbb{R}^{d'} ; W\boldsymbol{x} = \boldsymbol{0}\}.
\end{align}
This can be expressed using the singular value decomposition (SVD) of $W$;
\begin{align}
W &= U \Sigma V^T, \label{eq:svd_normal}
\end{align}
where, $\tilde{d}=\min(d, d')$, $U \in \mathbb{R}^{d \times \tilde{d}}$ is an orthogonal matrix consisting of left singular vectors, $\Sigma \in \mathbb{R}^{\tilde{d} \times  \tilde{d}}$ is a diagonal matrix with singular values $\sigma_1, \ldots, \sigma_{\tilde{d}}$ as diagonal components (sorted in descending order), and $V^T \in \mathbb{R}^{\tilde{d} \times d'}$ is an orthogonal matrix consisting of right singular vectors. 
If the least $s$ singular values $\sigma_{\tilde{d}-s+1}, \ldots, \sigma_{\tilde{d}}$ are zero, the column vectors of $U$ and the row vectors of $V^T$ corresponding to these parts can be separated as follows;
\begin{align}
W
&= 
\begin{bmatrix} U_m & U_n \end{bmatrix}
\begin{bmatrix}
\Sigma_m & 0 \\
0 & 0
\end{bmatrix}
\begin{bmatrix} V_m^T \\ V_n^T \end{bmatrix},
\label{eq:svd_decomposed}
\end{align}
where $\Sigma_m \in \mathbb{R}^{\tilde{d}-s \times \tilde{d}-s}$ is a diagonal matrix with nonzero singular values $\sigma_1, \ldots, \sigma_{\tilde{d}-s}$ on the diagonal, 
$U_m \in \mathbb{R}^{d \times \tilde{d}-s}$ is the basis for $\mathrm{col}(W)$, the column space of $W$, 
and $U_n \in \mathbb{R}^{d \times s}$ is a basis for a subspace of
$\mathrm{Ker}(W^T)$, the left null space of $W$.
Note that $U_n \subset \mathrm{Ker}(W^T)$ when $d < d'$,
however, we do not use the rest subspace $\mathrm{Ker}(W^T) \backslash U_n$ for simplicity, and we consider $U_n$ as the left null space in this paper.
We will investigate the use of the full left null space $\mathrm{Ker}(W^T) = I - U_m U_m^T$ in the future.

\subsubsection{Truncated SVD}

In our case, $W$ is the weight of the pre-trained model
and is full rank (\ie, $s=0$) in general, and therefore all singular values, $\sigma_1, \dots, \sigma_{\tilde{d}}$, are nonzero. 
Therefore, we truncate singular values smaller than a certain threshold and treat them as zero, as in truncated SVD \cite{tensor_book2021}.
In this work, we set a threshold for the cumulative contribution rate $C_k$ of the squared singular values;
\begin{equation}
\quad C_k = \frac{\sum_{i=1}^k \sigma_i^2}{\sum_{j=1}^{\tilde{d}} \sigma_j^2},
\quad k=1,2,\ldots, \tilde{d},
\label{eq:contribution_and_cumulative}
\end{equation}
and take the smallest $k$ for which $C_k$ exceeds the predefined ratio (95\%)
as the dimension of the column space.

\subsection{Constraints of LoRAs on subspaces}
\label{sec:Constraints of LoRAs on subspaces}

\begin{figure}[t]
    \centering
    \includegraphics[width=1\linewidth]{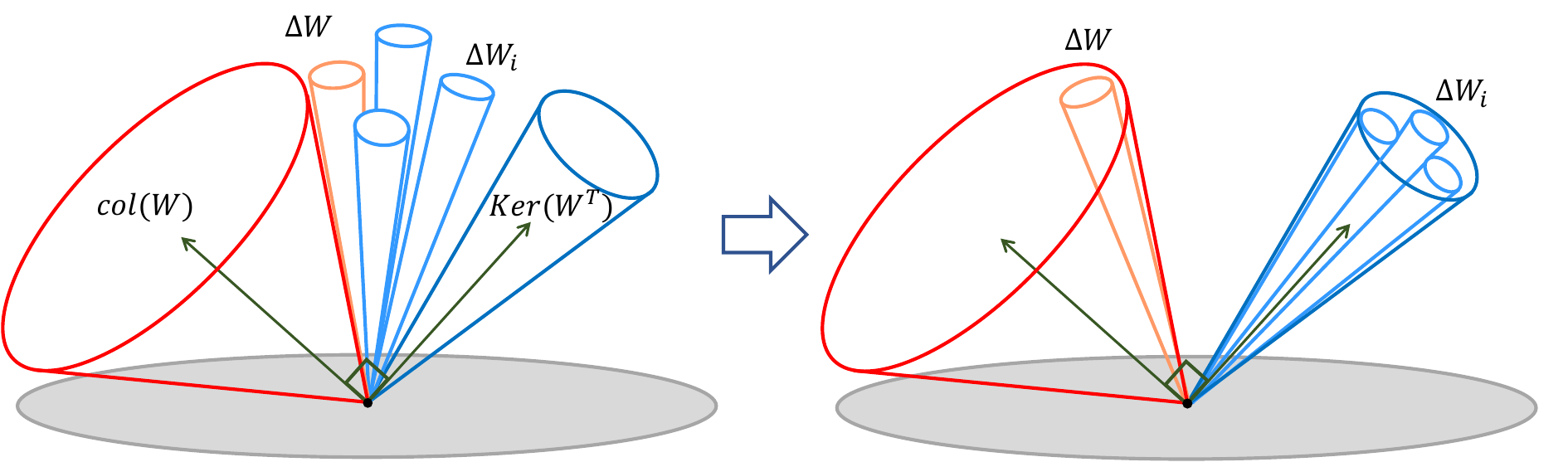}
    \caption{
    The proposed subspace constraints. The shared LoRA $\Delta W$ is constrained to the column space $\mathrm{col}(W)$ (red) and the domain-specific LoRAs $\Delta W_i$ to the left null space $\mathrm{Ker}(W^T)$ (blue).
    }
    \label{fig:bunri2}
\end{figure}

One of the core ideas of the proposed method is to restrict the shared LoRA $\Delta W$ to $\mathrm{col}(W)$ and the domain-specific LoRA $\Delta W_i$ to $\mathrm{Ker}(W^T)$, as shown in Figure \ref{fig:bunri2}. 
Since the important part of the information of $W$ is contained in the column space, we can interpret that the left null space of $W$ is not important to the pre-trained model.
Since the left null space and the column space are orthogonal to each other, we can efficiently learn different information by training different types of LoRAs in different subspaces.
More specifically, if we restrict domain-specific LoRA to the left null space of $W$ and shared LoRA to the column space of $W$, it would be expected to train domain-specific LoRA without affecting the features obtained from $W$, and enhance the knowledge contained in $W$ by training the shared LoRA to learn important features common across all domains.

The proposed method achieves this constraint by adding projection layers
as follows;
\begin{equation}
h = Wx + P_m \Delta W x + P_n \Delta W_i x,
\label{eq:lora_application}
\end{equation}
where $x$ and $h$ represent the input and output features,
and, the projection matrices $P_m = U_m U_m^T$ and $P_n = U_n U_n^T$ project features $x$ onto the column space $\mathrm{col}(W)$ and the left null space $\mathrm{Ker}(W^T)$, respectively.

The above projection is for the features, not for the LoRA weights.
However, this does not pose an issue.
The gradient of the LoRA weights, $\Delta W, \Delta W_i$, with respect to the loss function $L$ is calculated analytically as follows;
\begin{align}
  \frac{\partial L}{\partial \Delta W} 
  &= P_m^{T} \frac{\partial L}{\partial h} x^{T} \\
  \frac{\partial L}{\partial \Delta W_i} 
  &= P_n^{T} \frac{\partial L}{\partial h} x^{T}.
\label{eq:backprop}
\end{align}
Since $P_m^{T} = P_m$ and $P_n^{T} = P_n$ hold for the projection matrices,
each gradient vector is projected into the corresponding subspace,
that is, the gradient updates of the LoRA weights are only performed within each subspace. 

In the above update, only the components in the corresponding subspace are updated.
For example in the case of domain-specific LoRA, the gradient updates components of the left null space only, therefore the components in the column space of the initial values will remain unchanged.
To solve this situation, the proposed method projects the columns of randomly initialized weights $\Delta W, \Delta W_i$ into the corresponding subspaces once at the beginning of training.
This procedure effectively removes unnecessary components that exist in the subspace that do not correspond to the weights.

\subsection{Separating domain-specific LoRAs}
\label{sec:Separating domain-specific LoRAs}

The above projection enables explicit separation of shared and domain-specific LoRAs. However, it does not separate domain-specific LoRAs, that is, $\Delta W_i$ and $\Delta W_j$ for different domains ($i$ and $j$).
Therefore, we introduce losses to enforce the separation between domain-specific LoRAs, ensuring that each LoRA for each domain leans differently.
This approach was inspired by TesNet \cite{Wang_ICCV2021_TesNet}, 
which separates subspaces of different classes.
However, in our case, we separate subspaces of different domains.

Since a LoRA weight $\Delta W_i$ is composed of two matrices such as $\Delta W_i = B_i A_i^T$, 
$B_i$ is considered the corresponding domain-specific subspace. 
Here, we use the following requirements \cite{Wang_ICCV2021_TesNet};
$B_i$ has an orthogonal basis, 
and $B_i$ and $B_j$ ($i\neq j$) are as far as possible.

The first requirement of orthonormality is imposed by the following loss;
\begin{align}
    \mathcal{L}_{\mathrm{orth}} =
    \sum_{i=1}^{D} \left\| B_i^\top B_i - I_r \right\|_F^2,
    \label{eq:orth}
\end{align}
where $|\cdot|_F^2$ is the Frobenius norm, 
$I_r$ is the identity matrix of dimension $r$,
and $D$ represents the total number of domains.
By minimizing this loss, 
the correlation between each column vector of $B_i$ is suppressed, 
and each basis becomes as orthogonal to each other as possible.

The second requirement of subspace separation is achieved by the loss;
\begin{equation}
    \mathcal{L}_{\mathrm{ss}} = -\frac{1}{\sqrt{2}} 
    \sum_{i=1}^{D-1} \sum_{j=i+1}^{D} 
    \left\| B_i B_i^\top - B_j B_j^\top \right\|_F, 
    \label{eq:subspace}
\end{equation}
where $B_{d_1}$ and $B_{d_2}$ represent the basis matrices corresponding to domains $d_1$ and $d_2$, respectively.

The final loss for the domain-specific LoRAs is the sum of them;
\begin{equation}
\mathcal{L}_{\mathrm{total}} = \mathcal{L}_{\mathrm{CE}} + \lambda_1 \mathcal{L}_{\mathrm{orth}} + \lambda_2 \mathcal{L}_{\mathrm{ss}},
\end{equation}
where $\lambda_1$ and $\lambda_2$ are hyperparameters to adjust the balance of each term, and $\mathcal{L}_{\mathrm{CE}}$ is the cross-entropy loss used in a normal training.

\section{Experimental Results}

We show experiments to evaluate the effect of the constraings of the shared and domain-specific LoRAs to different subspaces, and separations between domain-specific LoRAs.

\begin{figure*}[t]
    \centering
    \begin{minipage}{0.8\linewidth}
        \centering
        \includegraphics[width=\linewidth]{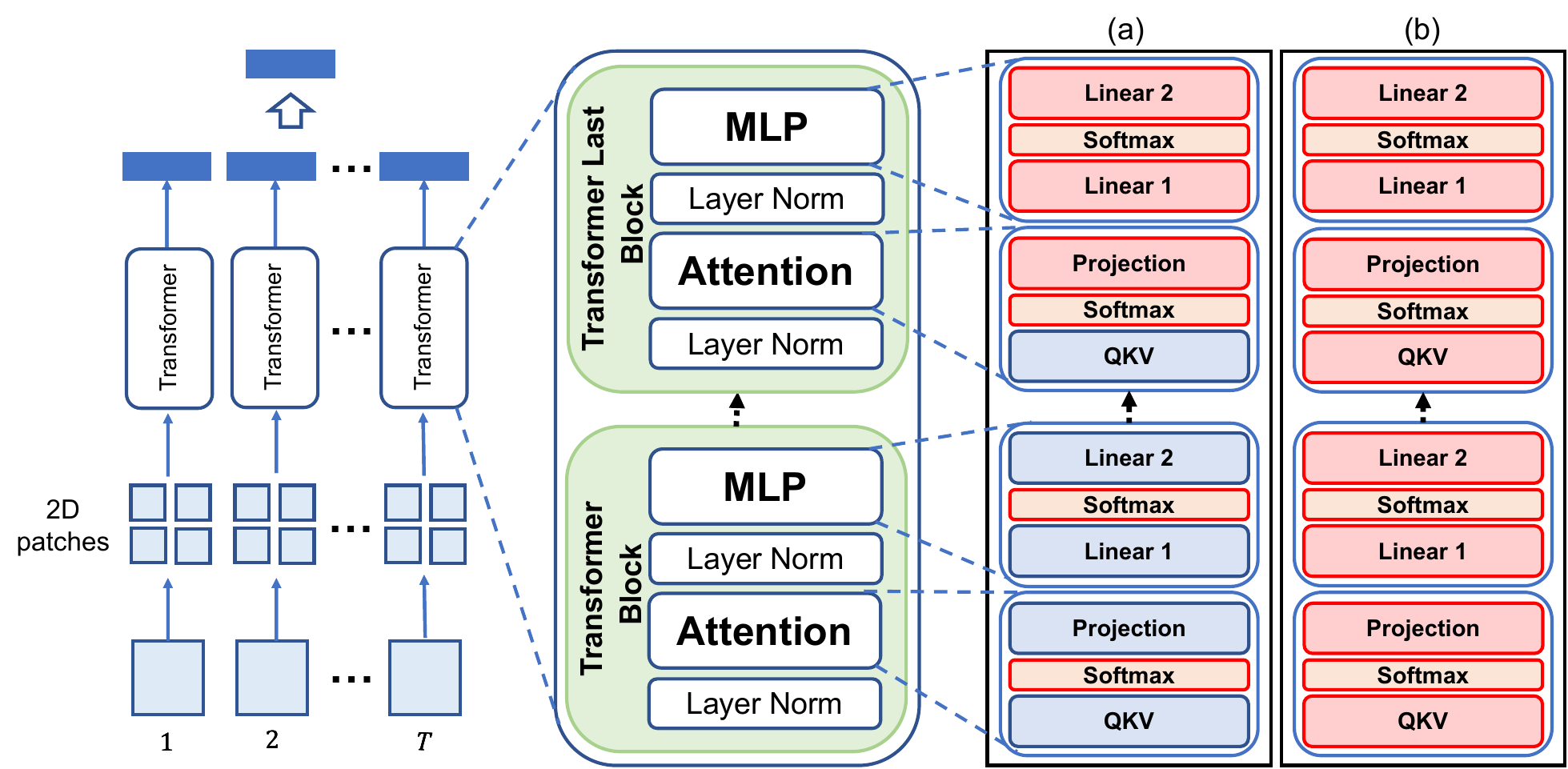}
        \label{fig:main_image}
    \end{minipage}%
    \hfill
    \begin{minipage}{0.15\linewidth}
        \centering
        \includegraphics[width=\linewidth]{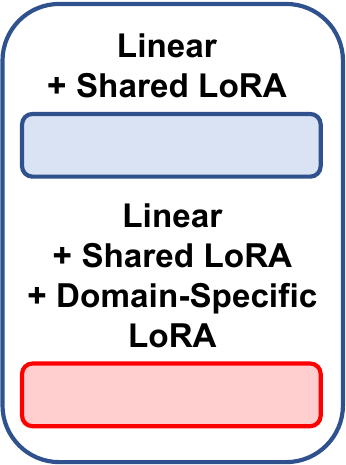}
    \end{minipage}
    \caption{
    Two structures used in the experiments.
    (a) The upper-heavy structure similar to
    \cite{Agiza_CVPR2024_MTLoRA};
    both LoRAs are used only at the last layers in a Transformer block, and other layers have shared LoRA only.
    (b) The all-flat structure in which all layers have both LoRAs.
    }
    \label{fig:model}
\end{figure*}

\subsection{Setting}

\subsubsection{Datasets and models}

As mentioned in the introduction,
we focus on the task of action recognition.
For the experiments, we used
three common action recognition datasets;
Kinetics400 \cite{kay_arXiv2017_kinetics400}, UCF101 \cite{Soomro_arXiv2012_UCF101}, and HMDB51 \cite{Kuehne_ICCV2011_HMDB51}.
Kinetics400 \cite{kay_arXiv2017_kinetics400} is a dataset consisting of 219k training videos and 18k validation videos. Each video was collected from YouTube, and the corresponding part of each category was cropped at a length of 10 seconds.
UCF101 \cite{Soomro_arXiv2012_UCF101} is a dataset of 101 classes of human actions, consisting of a training set of 9.5k videos and a validation set of 3.5k videos. Each video was collected from YouTube, and the average video length is 7.21 seconds.
HMDB51 \cite{Kuehne_ICCV2011_HMDB51} is a dataset of 51 classes of human actions, consisting of a training set of 3.6k videos and a validation set of 1.5k videos. Each video was collected from movies, the web, YouTube, etc., and the average video length is 3.15 seconds.

For all datasets, we set the learning rate at 1e-4, the batch size at 8, the number of training iterations at 50000, and we used AdamW \cite{Loshchilov_ICLR2019_AdamW} with
a MultiStepLR to schedule the learning rate. 
The rank $r$ of LoRA was fixed at 32, and the threshold of the cumulative contribution rate was fixed at 0.95. $\lambda_1$ and $\lambda_2$, were set at 1 and 1e-7, respectively.

As the model to predict the action category,
we used a model with multiple heads for each domain
with a single body of Vision Transformer (ViT) \cite{Dosovitskiy_ICLR2021_ViT_Vision_transformer}
pre-trained on ImageNet21K \cite{ridnik_arXiv2021_imagenet21k_pretraining_masses}.
We adopted the late-fusion structure, which is a common
approach for applying ViT to action recognition, to average the features of each frame of the input video.
For each linear layer $W$ of this pre-trained model,
SVD is performed to obtain
the column space $\mathrm{col}(W)$ and the left null space $\mathrm{Ker}(W^T)$.
Typically, the dimensionality $s$ of the left null space is roughly 50\% of
the dimension $d$ of the entire space, with a range of 20 to 80\%.

\subsubsection{LoRAs}

We follow MTLoRA \cite{Agiza_CVPR2024_MTLoRA}
to decide to which linear layers LoRAs are inserted;
the attention part (query, key, and value) of each layer of the Transformer, the projection layer and the two-layer MLP (Linear1 and Linear2). 
The original pre-trained ViT parameters are frozen,
while the LoRA weights are fine-tuned.

When inserting LoRA into a particular linear layer, 
there are two choices: use only shared LoRA, or use both shared LoRA and domain-specific LoRA.
Based on these choices, 
we will compare the following two structures (Figure \ref{fig:model}).
The first is similar to MTLoRA \cite{Agiza_CVPR2024_MTLoRA},
called the \emph{upper-heavy} type,
which uses both LoRAs in the projection layer and MLP in the last block of the Transformer and uses shared LoRA only in the other layers. 
The second structure uses both LoRAs in all linear layers, called the \emph{all-flat} type.
Note that
the proposed method is performed only in the layer with both LoRAs because
there are no domain-specific LoRAs in such layers.

\begin{table}[t]
\centering
\caption{
Performance comparison of LoRA integration methods with and without subspace constraints (sc) and domain-specific LoRAs separation (ds)
}

    \begin{tabular}{c|c|c|c|c|c}
    structure & sc & ds & UCF & K400 & HMDB  \\ \hline
    \multirow{3}{*}{\shortstack{upper\\ heavy}} 
        &  & & 92.69 & \textbf{58.10} & 66.05 \\  
        & \checkmark &  & \textbf{92.88} & 57.81 & 66.05 \\  
        & \checkmark & \checkmark & 91.74 & 57.77 & \textbf{66.45} \\ \hline
    \multirow{3}{*}{\shortstack{all\\ flat}}  
        &  &  & 90.52 & \textbf{61.05} & \textbf{64.01} \\  
        & \checkmark &  & 90.39 & 58.32 & 62.30 \\  
        & \checkmark & \checkmark& \textbf{91.31} & 57.97 & 63.16 \\  
    \end{tabular}
    

\label{tab:main_result}

\end{table}

\subsection{Results}

\subsubsection{Comparison of structures}

First, we compared two structures
in Figure \ref{fig:model} with and without the subspace constraints (sc) of LoRAs and the domain-specific LoRAs separation (ds).
The results are shown in Table \ref{tab:main_result}.

Without both subspace constraints and domain-specific LoRA separation,
the all-flat structure with both LoRAs in all layers
performs better for Kinetics only.
This might be due to the increased number of LoRA parameters, leading to a better performance only for the relatively difficult Kinetics, while causing overfitting for UCF and HMDB.
With the proposed subspace constraints and the separation, the performance improved slightly for UCF and HMDB with the upper-heavy structure,
which is a promising result, but further exploration of the structure is necessary.
In the following experiments and analysis, we use the upper-heavy structure.

\subsubsection{Rank $r$ of LoRA}

The rank $r$ of the matrices $B, B_i$ in LoRA is an important hyperparameter that fundamentally controls the learning capacity of LoRAs.
Table \ref{tab:lora_r_experiment} shows the performance across different rank values $r$ of LoRAs, only with subspace constraints. In general, performance improves as $r$ increases, although excessively large values can degrade performance.
Specifically, we observed the best performance on UCF101 and Kinetics400 when setting $r=32$, and on HMDB51 $r=16$. Based on this observation, we set $r=32$ for other experiments.

\begin{table}[t]
\centering
\caption{
Performances with different rank $r$ of LoRA.
}
\begin{tabular}{c|c|c|c}
$r$ & UCF   & K400  & HMDB  \\ \hline
4            & 90.89        & 54.36         & 63.42        \\ 
8            & 91.92          & 56.16          & 65.13          \\ 
16           & 91.92          & 57.10          & \textbf{67.37}          \\ 
32           & \textbf{92.88}         & \textbf{57.81}          & 66.05          \\ 
64           & 92.29          & 57.59          & 66.18          \\ 
\end{tabular}
\label{tab:lora_r_experiment}
\end{table}

\subsubsection{Threshold to cumulative contribution ratio}

The proposed method needs the threshold for the cumulative contribution ratio $C_k$ to determine the dimensionality $s$ of the left null space. 
Table \ref{tab:lora_c_experiment}
shows the performance across different thresholds with the subspace constraints only.
Depending on the dataset, 
the threshold was shown to have a significant impact on performance. 
For UCF101 and Kinetics400, the threshold values had little impact on performance,
but for HMDB51, the highest performance was achieved with the smallest threshold, which leads to a better performance by more than one point.
This observation needs more investigation because, in general, a subspace is well approximated with a higher threshold.
We used 0.95 as the threshold value for other experiments,
but a better solution for determining the threshold will be explored in future.

\begin{table}[t]
\centering
\caption{
Performances with different threshold for cumulative contribution ratio.
}
\begin{tabular}{c|c|c|c}
threshold & UCF   & K400  & HMDB  \\ \hline
0.80            & 92.24          & 57.62          & \textbf{67.50}          \\ 
0.90            & 92.37  & 57.52          & 66.45          \\ 
0.95           & \textbf{92.88}         & \textbf{57.81}    & 66.05          \\ 
0.99           & 92.14          & 57.57          & 66.05          \\ 
\end{tabular}
\label{tab:lora_c_experiment}
\end{table}

\subsubsection{Analysis on effective dimensions of LoRAs}

The proposed method restricts LoRA weights of dimension $r=32$ to the corresponding subspaces instead of the entire $d$-dimensional space.
Therefore, this affects the effective dimensions of the LoRAs,
and they can represent their own subspaces with dimensions smaller than $r$ because each domain-specific LoRA can focus on their own domain.

\begin{figure*}[t]
    \centering

        \begin{subfigure}{0.3\linewidth}
            \centering
            \includegraphics[width=\linewidth]{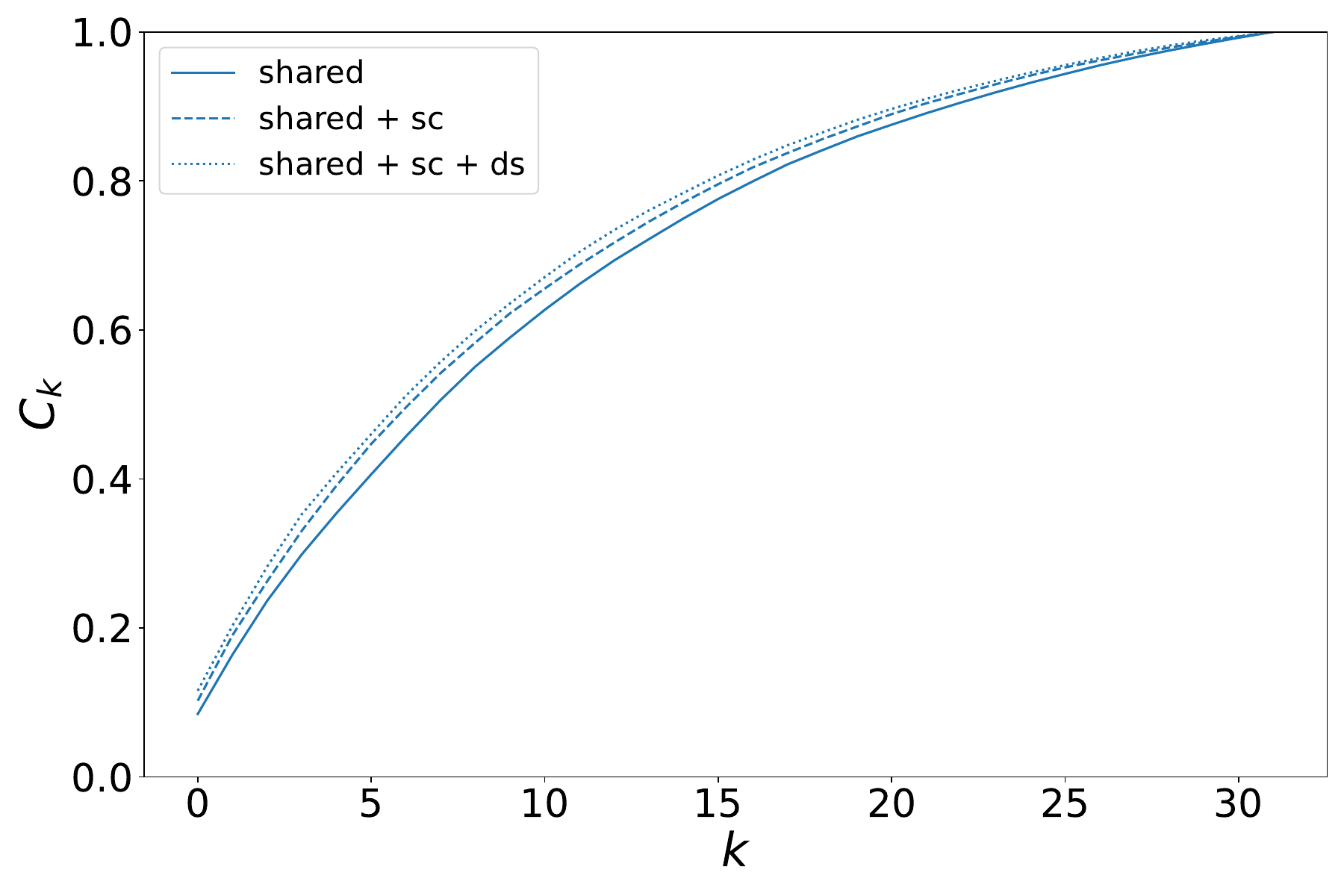}
            \caption{shared LoRA (projection layer)}
            \label{fig:projection}
        \end{subfigure}
        \hfill
        \begin{subfigure}{0.3\linewidth}
            \centering
            \includegraphics[width=\linewidth]{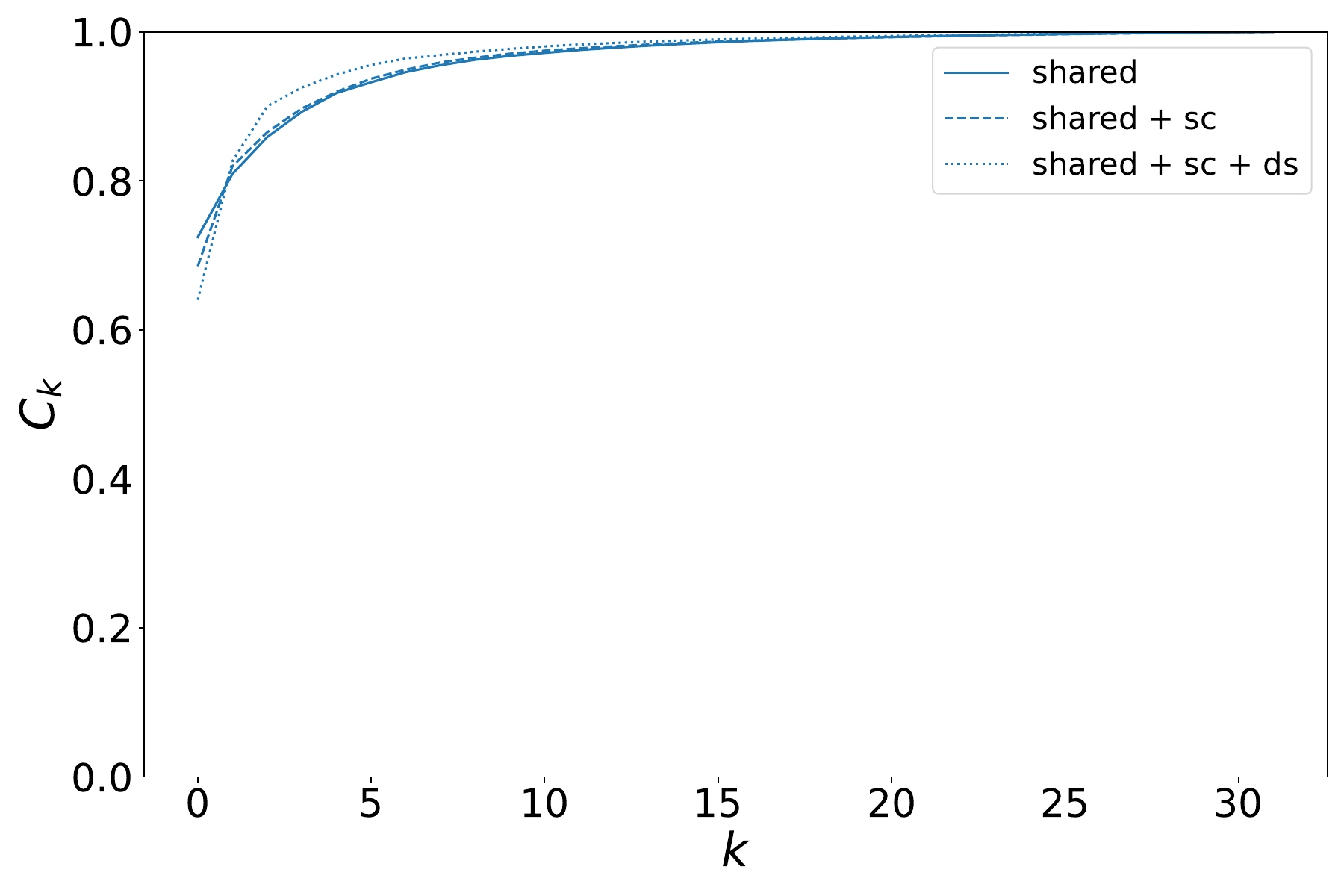}
            \caption{shared LoRA (MLP Linear1)}
            \label{fig:Linear1}
        \end{subfigure}
        \hfill
        \begin{subfigure}{0.3\linewidth}
            \centering
            \includegraphics[width=\linewidth]{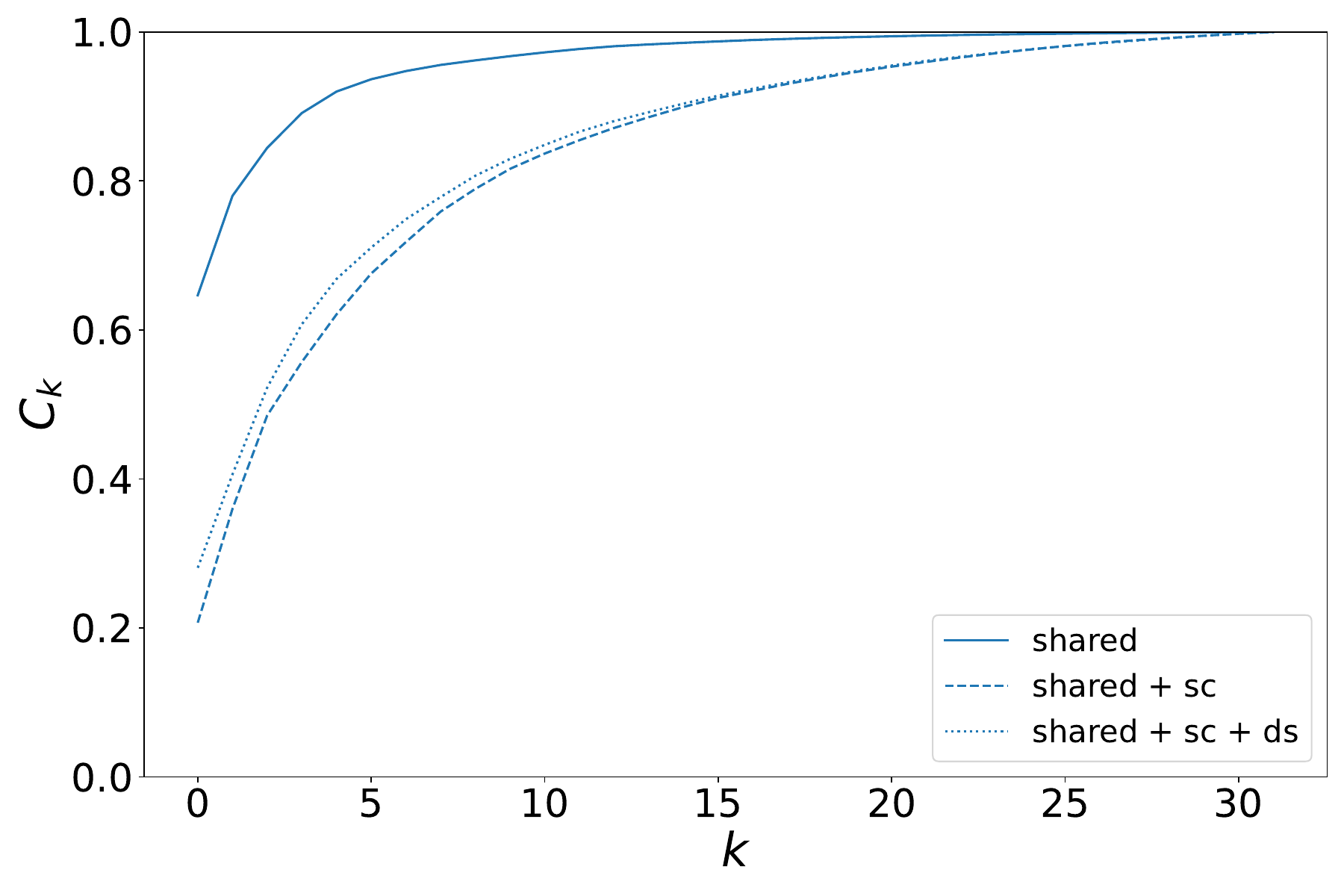}
            \caption{shared LoRA (MLP Linear2)}
            \label{fig:Linear2}
        \end{subfigure}

        \begin{subfigure}{0.3\linewidth}
            \centering
            \includegraphics[width=\linewidth]{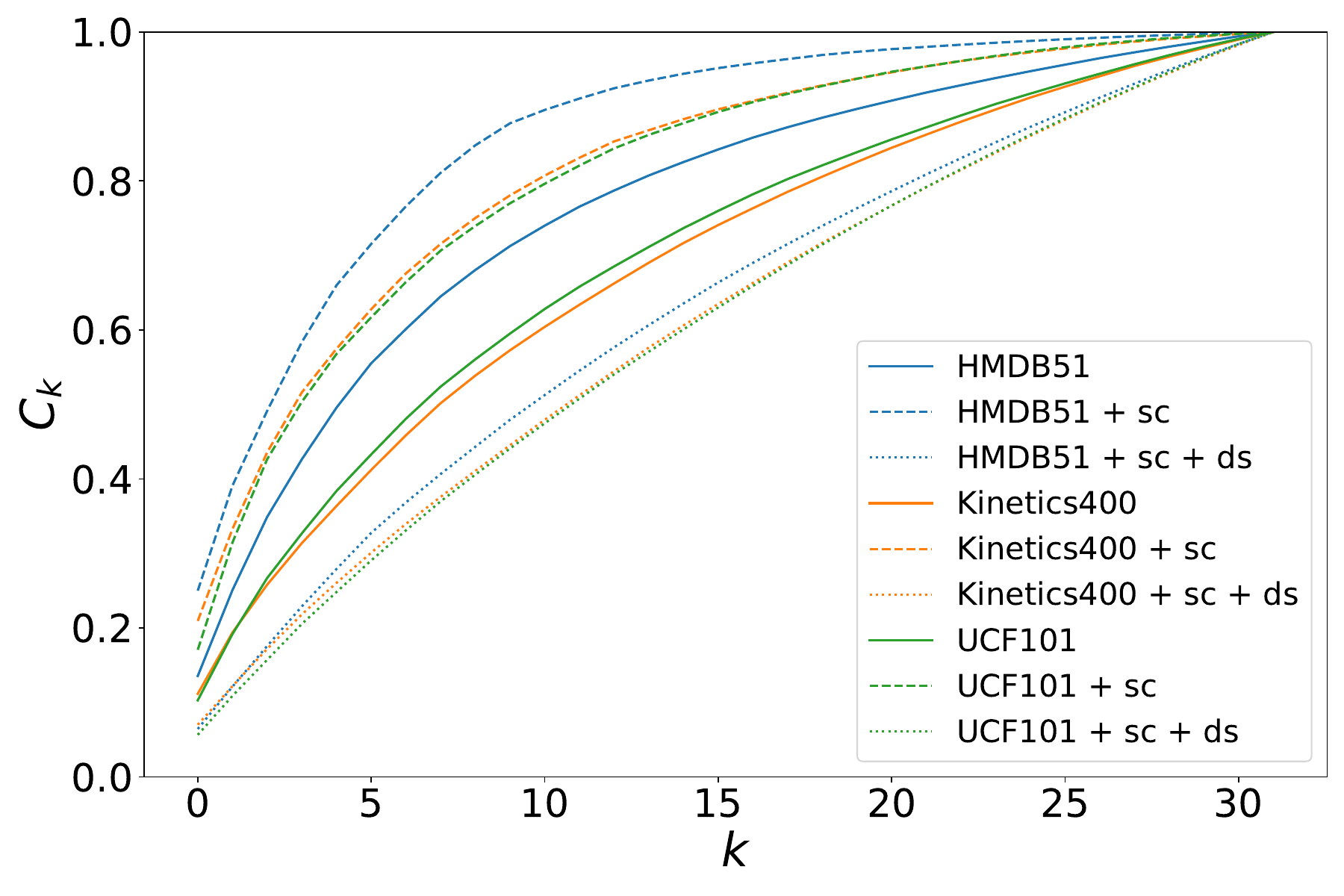}
            \caption{domain-specific LoRA (projection layer)}
            \label{fig:specific_projection}
        \end{subfigure}
        \hfill
        \begin{subfigure}{0.3\linewidth}
            \centering
            \includegraphics[width=\linewidth]{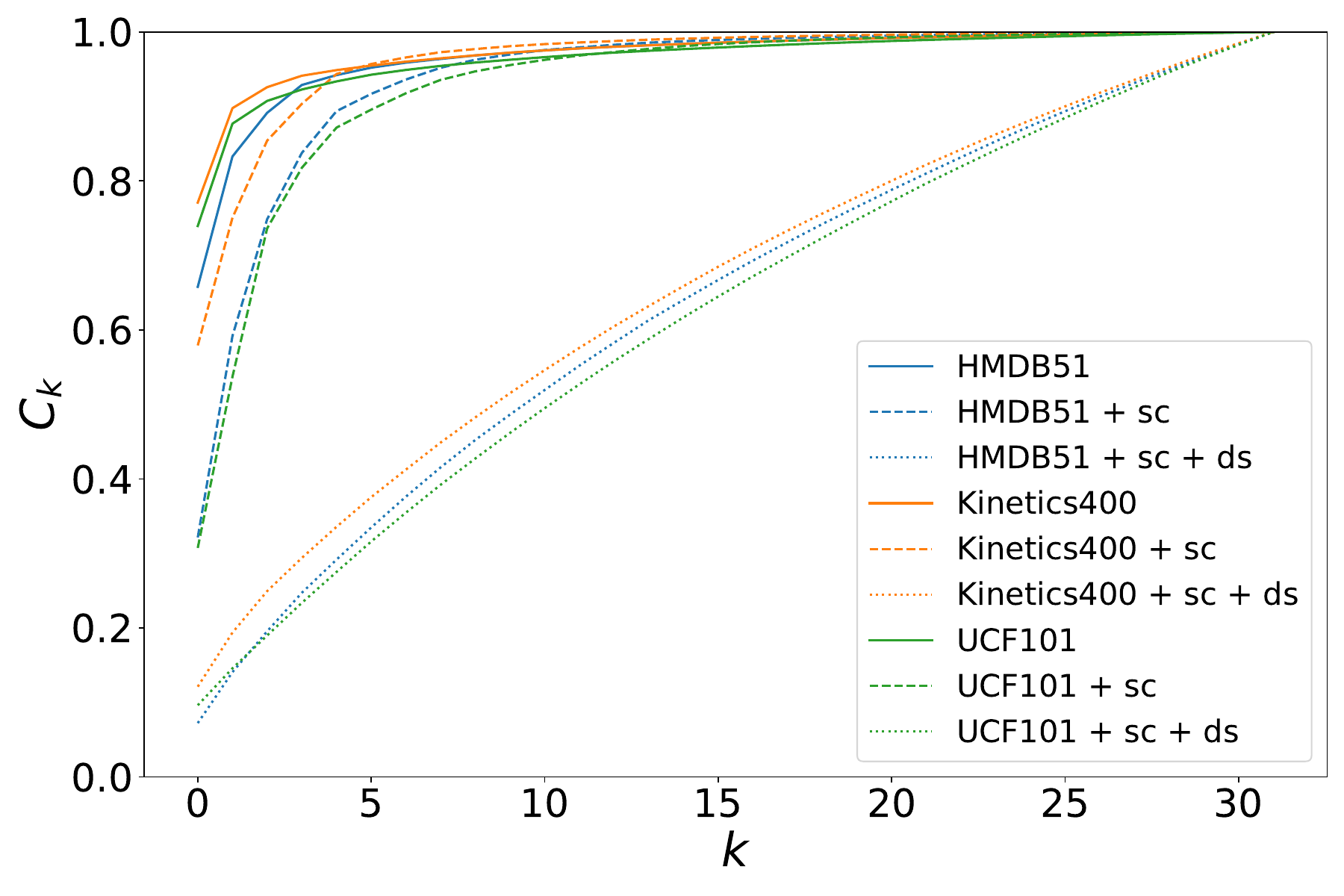}
            \caption{domain-specific LoRA (MLP Linear1)}
            \label{fig:specific_Linear1}
        \end{subfigure}
        \hfill
        \begin{subfigure}{0.3\linewidth}
            \centering
            \includegraphics[width=\linewidth]{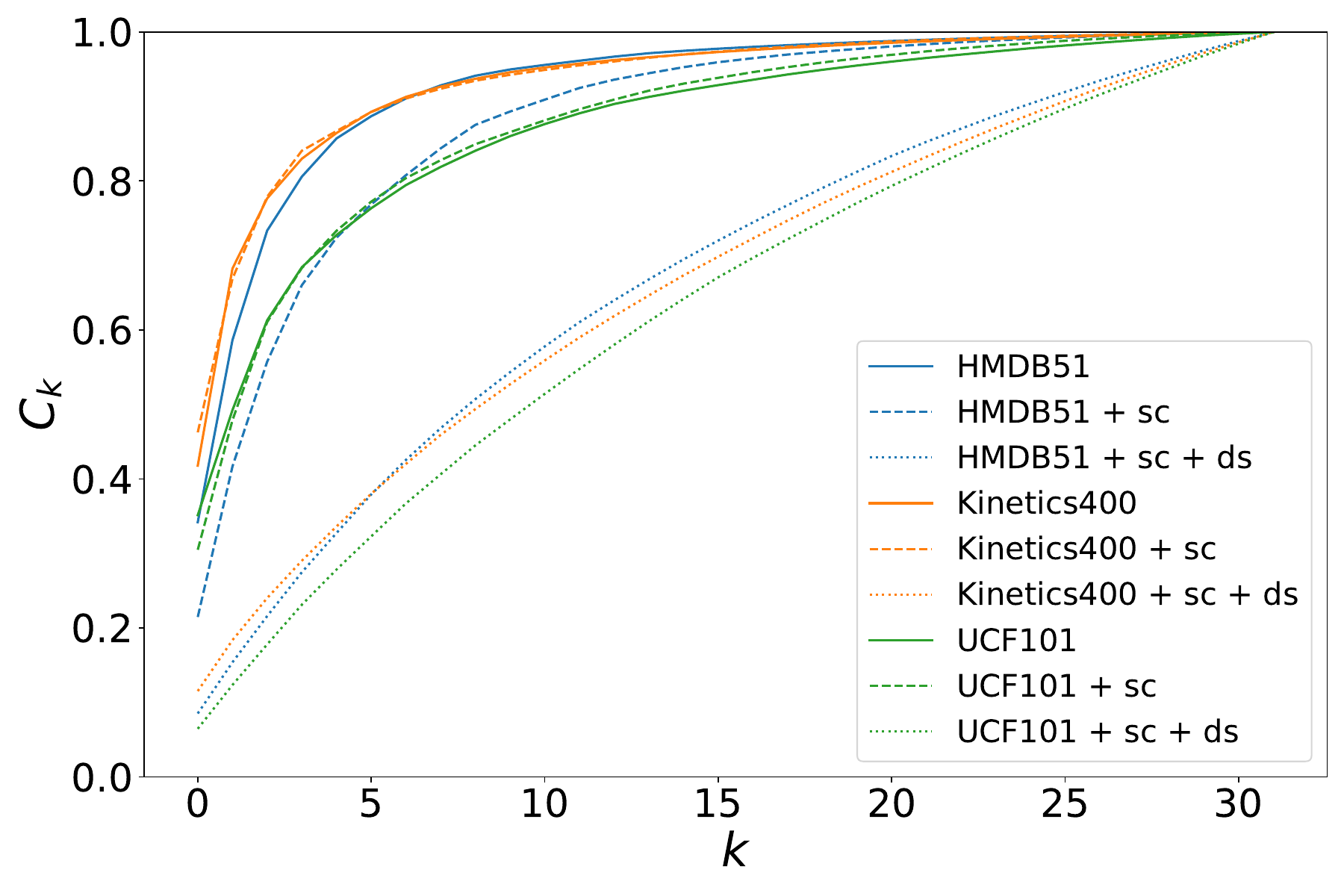}
            \caption{domain-specific LoRA (MLP Linear2)}
            \label{fig:specific_Linear2}
        \end{subfigure}

    \caption{
    Comparisons of the cumulative contribution ratio of the LoRA weights for
    (upper row) shared LoRA and (bottom row) domain-specific LoRA
    in the different layers (projection layer and two MLP layers,
    see Fig. \ref{fig:model}).
    The horizontal axis shows the column number $k$ of the left singular vector of SVD, and the vertical axis is the cumulative contribution rate up to $k$-th column. 
    }
    \label{fig:compare}
\end{figure*}

Figure \ref{fig:compare} shows the cumulative contribution ratio of
the LoRA weights. More specifically, we applied SVD to the LoRA matrices
then used Eq.\eqref{eq:contribution_and_cumulative} to calculate $C_k$ for $k=1,\ldots,r$.
We used the upper-heavy structure in these experiments, where both shared and domain-specific LoRAs were added to only three layers in the transformer's final block: the projection layer, linear1, and linear2 layers in MLP.
The top three plots of Fig. \ref{fig:compare} show the curves for shared LoRA, while the bottom three plots show the curves for domain-specific LoRAs across three domains. The curve colors show three domains (UCF100, Kinetics400, and HMDB51), with different line styles for different configurations: solid lines for the case with no constraints, dashed lines for the subspace constraint (sc) only, and dotted lines for both the subspace constraint and domain-specific LoRA separation (ds).

With a threshold of 95\% for the cumulative contribution ratio, the dimensions of column and left null spaces of the three layers are as follows: for the projection layer, $\operatorname{dim}(\mathrm{col}(W))=166$ and $\operatorname{dim}(\mathrm{Ker}(W^T))=602$; for the linear1 layer, $\operatorname{dim}(\mathrm{col}(W))=146$ and $\operatorname{dim}(\mathrm{Ker}(W^T))=622$; and for the linear2 layer, $\operatorname{dim}(\mathrm{col}(W))=151$ and $\operatorname{dim}(\mathrm{Ker}(W^T))=617$. Consequently, in each layer, the 32-dimensional shared LoRA was obtained within a column space of roughly 150 dimensions, while the three domain-specific LoRAs each use 32 dimensions within about a left null space of about 600 dimensions.

For the linear1 and linear2 layers, the curves increase rapidly, suggesting that, as expected, the LoRAs concentrate in a subspace smaller than $r$.
When using the subspace constraint, the domain-specific LoRA of the projection layer shows a curve shift toward the upper left, indicating a reduction in effective dimensions. 
For other layers and LoRAs, the subspace constraint has minimal impact on the effective dimensions. However, adding domain-specific LoRA separation causes the curves to shift toward the lower right and become linear, showing that the basis fully utilizes all 32 dimensions. This is because the null space dimension of 600 is sufficiently large compared to the 96 dimensions needed for the three domain-specific LoRAs, resulting in a less restrictive constraint. As the number of domains increases, domain-specific LoRAs are expected to have smaller dimensions.

\section{Conclusion} 

In this paper, we proposed a method for explicitly separating shared and domain-specific LoRAs by constraining them to the column and left null spaces, and further separating domain-specific LoRAs.
In experiments using three action recognition datasets with a frozen model trained on an image recognition dataset,
we investigated how subspace constraints and the separation of domain-specific LoRAs affect performance. 
Although the results varied according to the hyperparameters and further investigation is needed, we obtained promising results in some cases.
Future work includes further exhaustive experiments,
addition of other domains and datasets, and investigation of
other constraints.

\section*{\uppercase{Acknowledgments}}
This manuscript was written with the help of the following AI-based tools;
Notion and Writefull for editing and grammar enhancement,
chatGPT for rephrasing and equation conversion, and 
DeepL for translation.
Y. Takama thanks S. Shimizu for fruitful discussions and the initial codebase of the project.
This work was supported in part by JSPS KAKENHI Grant Number JP22K12090 and 25K03138.

{
    \small
    \bibliographystyle{ieeenat_fullname}
    \bibliography{mybib,all}

\begin{thebibliography}{43}
\providecommand{\natexlab}[1]{#1}
\providecommand{\url}[1]{\texttt{#1}}
\expandafter\ifx\csname urlstyle\endcsname\relax
  \providecommand{\doi}[1]{doi: #1}\else
  \providecommand{\doi}{doi: \begingroup \urlstyle{rm}\Url}\fi

\bibitem[Agiza et~al.(2024)Agiza, Neseem, and Reda]{Agiza_CVPR2024_MTLoRA}
Ahmed Agiza, Marina Neseem, and Sherief Reda.
\newblock {MTLoRA}: Low-rank adaptation approach for efficient multi-task learning.
\newblock In \emph{Proceedings of the IEEE/CVF Conference on Computer Vision and Pattern Recognition (CVPR)}, pages 16196--16205, 2024.

\bibitem[Crawshaw(2020)]{Crawshaw_MTL_survey_2020}
Michael Crawshaw.
\newblock Multi-task learning with deep neural networks: {A} survey.
\newblock \emph{CoRR}, abs/2009.09796, 2020.

\bibitem[Dettmers et~al.(2023)Dettmers, Pagnoni, Holtzman, and Zettlemoyer]{Dettmers_QLoRA_2023}
Tim Dettmers, Artidoro Pagnoni, Ari Holtzman, and Luke Zettlemoyer.
\newblock {QLoRA}: Efficient finetuning of quantized llms.
\newblock In \emph{Advances in Neural Information Processing Systems}, pages 10088--10115, 2023.

\bibitem[Devlin et~al.(2019)Devlin, Chang, Lee, and Toutanova]{Devlin-ACL2019-BERT}
Jacob Devlin, Ming-Wei Chang, Kenton Lee, and Kristina Toutanova.
\newblock {BERT}: Pre-training of deep bidirectional transformers for language understanding.
\newblock In \emph{Proceedings of the 2019 Conference of the North {A}merican Chapter of the Association for Computational Linguistics: Human Language Technologies, Volume 1 (Long and Short Papers)}, pages 4171--4186, Minneapolis, Minnesota, 2019. Association for Computational Linguistics.

\bibitem[Dosovitskiy et~al.(2021)Dosovitskiy, Beyer, Kolesnikov, Weissenborn, Zhai, Unterthiner, Dehghani, Minderer, Heigold, Gelly, Uszkoreit, and Houlsby]{Dosovitskiy_ICLR2021_ViT_Vision_transformer}
Alexey Dosovitskiy, Lucas Beyer, Alexander Kolesnikov, Dirk Weissenborn, Xiaohua Zhai, Thomas Unterthiner, Mostafa Dehghani, Matthias Minderer, Georg Heigold, Sylvain Gelly, Jakob Uszkoreit, and Neil Houlsby.
\newblock An image is worth 16x16 words: Transformers for image recognition at scale.
\newblock In \emph{International Conference on Learning Representations}, 2021.

\bibitem[Girdhar et~al.(2022)Girdhar, Singh, Ravi, van~der Maaten, Joulin, and Misra]{Girdhar_Omnivore_CVPR2022}
Rohit Girdhar, Mannat Singh, Nikhila Ravi, Laurens van~der Maaten, Armand Joulin, and Ishan Misra.
\newblock Omnivore: A single model for many visual modalities.
\newblock In \emph{Proceedings of the IEEE/CVF Conference on Computer Vision and Pattern Recognition (CVPR)}, pages 16102--16112, 2022.

\bibitem[Han et~al.(2024)Han, Gao, Liu, Zhang, and Zhang]{Han_PEFT_survey_2024}
Zeyu Han, Chao Gao, Jinyang Liu, Jeff Zhang, and Sai~Qian Zhang.
\newblock Parameter-efficient fine-tuning for large models: A comprehensive survey.
\newblock \emph{Transactions on Machine Learning Research}, 2024.

\bibitem[He et~al.(2016)He, Zhang, Ren, and Sun]{He_CVPR2016_ResNet}
Kaiming He, Xiangyu Zhang, Shaoqing Ren, and Jian Sun.
\newblock Deep residual learning for image recognition.
\newblock In \emph{Proceedings of the IEEE Conference on Computer Vision and Pattern Recognition (CVPR)}, pages 770--778, 2016.

\bibitem[Houlsby et~al.(2019)Houlsby, Giurgiu, Jastrzebski, Morrone, De~Laroussilhe, Gesmundo, Attariyan, and Gelly]{Houlsby_PMLR2019_Parameter-Efficient}
Neil Houlsby, Andrei Giurgiu, Stanislaw Jastrzebski, Bruna Morrone, Quentin De~Laroussilhe, Andrea Gesmundo, Mona Attariyan, and Sylvain Gelly.
\newblock Parameter-efficient transfer learning for {NLP}.
\newblock In \emph{Proceedings of the 36th International Conference on Machine Learning}, pages 2790--2799. PMLR, 2019.

\bibitem[Hu et~al.(2022)Hu, yelong shen, Wallis, Allen-Zhu, Li, Wang, Wang, and Chen]{hu_arXiv2021_lora_low_rank_adaptation_large}
Edward~J Hu, yelong shen, Phillip Wallis, Zeyuan Allen-Zhu, Yuanzhi Li, Shean Wang, Lu Wang, and Weizhu Chen.
\newblock {LoRA}: Low-rank adaptation of large language models.
\newblock In \emph{International Conference on Learning Representations}, 2022.

\bibitem[Karimi~Mahabadi et~al.(2021)Karimi~Mahabadi, Ruder, Dehghani, and Henderson]{Karimi_Hyperformer_ACL2021}
Rabeeh Karimi~Mahabadi, Sebastian Ruder, Mostafa Dehghani, and James Henderson.
\newblock Parameter-efficient {Multi}-task {Fine}-tuning for {Transformers} via {Shared} {Hypernetworks}.
\newblock In \emph{Proceedings of the 59th {Annual} {Meeting} of the {Association} for {Computational} {Linguistics} and the 11th {International} {Joint} {Conference} on {Natural} {Language} {Processing} ({Volume} 1: {Long} {Papers})}, pages 565--576, Online, 2021. Association for Computational Linguistics.

\bibitem[Kay et~al.(2017)Kay, Carreira, Simonyan, Zhang, Hillier, Vijayanarasimhan, Viola, Green, Back, Natsev, Suleyman, and Zisserman]{kay_arXiv2017_kinetics400}
Will Kay, Jo{\~{a}}o Carreira, Karen Simonyan, Brian Zhang, Chloe Hillier, Sudheendra Vijayanarasimhan, Fabio Viola, Tim Green, Trevor Back, Paul Natsev, Mustafa Suleyman, and Andrew Zisserman.
\newblock The kinetics human action video dataset.
\newblock \emph{CoRR}, abs/1705.06950, 2017.

\bibitem[Kong and Fu(2022)]{Kong_IJCV2022_Action_Recognition_Survey}
Yu Kong and Yun Fu.
\newblock Human action recognition and prediction: {A} survey.
\newblock \emph{International Journal of Computer Vision}, 130\penalty0 (5):\penalty0 1366--1401, 2022.

\bibitem[Kuehne et~al.(2011)Kuehne, Jhuang, Garrote, Poggio, and Serre]{Kuehne_ICCV2011_HMDB51}
Hildegard Kuehne, Hueihan Jhuang, Est{\'{\i}}baliz Garrote, Tomaso~A. Poggio, and Thomas Serre.
\newblock {HMDB:} {A} large video database for human motion recognition.
\newblock In \emph{{IEEE} International Conference on Computer Vision, {ICCV} 2011, Barcelona, Spain, November 6-13, 2011}, pages 2556--2563. {IEEE} Computer Society, 2011.

\bibitem[Li and Vasconcelos(2019{\natexlab{a}})]{Li_Efficient_Multi_Domain_CVPR2019}
Yunsheng Li and Nuno Vasconcelos.
\newblock Efficient multi-domain learning by covariance normalization.
\newblock In \emph{Proceedings of the IEEE/CVF Conference on Computer Vision and Pattern Recognition (CVPR)}, 2019{\natexlab{a}}.

\bibitem[Li and Vasconcelos(2019{\natexlab{b}})]{Li_REPAIR_CVPR2019}
Yi Li and Nuno Vasconcelos.
\newblock {REPAIR}: Removing representation bias by dataset resampling.
\newblock In \emph{Proceedings of the IEEE/CVF Conference on Computer Vision and Pattern Recognition (CVPR)}, 2019{\natexlab{b}}.

\bibitem[Li et~al.(2018)Li, Li, and Vasconcelos]{Ferrari_RESOUND_ECCV2018}
Yingwei Li, Yi Li, and Nuno Vasconcelos.
\newblock {RESOUND}: {Towards} {Action} {Recognition} {Without} {Representation} {Bias}.
\newblock In \emph{Computer {Vision} – {ECCV} 2018}, pages 520--535, 2018.

\bibitem[Liang et~al.(2022)Liang, Zhang, Zhang, and Shen]{Liang_Multi_Dataset_NeurIPS2022}
Junwei Liang, Enwei Zhang, Jun Zhang, and Chunhua Shen.
\newblock Multi-dataset {Training} of {Transformers} for {Robust} {Action} {Recognition}.
\newblock \emph{Advances in Neural Information Processing Systems}, 35:\penalty0 14475--14488, 2022.

\bibitem[Likhosherstov et~al.(2023)Likhosherstov, Arnab, Choromanski, Lucic, Tay, and Dehghani]{Likhosherstov_Polyvit_2021}
Valerii Likhosherstov, Anurag Arnab, Krzysztof~Marcin Choromanski, Mario Lucic, Yi Tay, and Mostafa Dehghani.
\newblock {PolyViT}: Co-training vision transformers on images, videos and audio.
\newblock \emph{Transactions on Machine Learning Research}, 2023.

\bibitem[Liu et~al.(2022)Liu, Ma, Tian, He, and Kira]{Liu_Polyhistor_NeurIPS2022}
Yen-Cheng Liu, Chih-Yao Ma, Junjiao Tian, Zijian He, and Zsolt Kira.
\newblock Polyhistor: {Parameter}-{Efficient} {Multi}-{Task} {Adaptation} for {Dense} {Vision} {Tasks}.
\newblock \emph{Advances in Neural Information Processing Systems}, 35:\penalty0 36889--36901, 2022.
\newblock Polyhistor.

\bibitem[Loshchilov and Hutter(2019)]{Loshchilov_ICLR2019_AdamW}
Ilya Loshchilov and Frank Hutter.
\newblock Decoupled weight decay regularization.
\newblock In \emph{International Conference on Learning Representations}, 2019.

\bibitem[Radford et~al.(2021)Radford, Kim, Hallacy, Ramesh, Goh, Agarwal, Sastry, Askell, Mishkin, Clark, Krueger, and Sutskever]{Radford_ICML2021_CLIP}
Alec Radford, Jong~Wook Kim, Chris Hallacy, Aditya Ramesh, Gabriel Goh, Sandhini Agarwal, Girish Sastry, Amanda Askell, Pamela Mishkin, Jack Clark, Gretchen Krueger, and Ilya Sutskever.
\newblock Learning transferable visual models from natural language supervision.
\newblock In \emph{Proceedings of the 38th International Conference on Machine Learning}, pages 8748--8763. PMLR, 2021.

\bibitem[Rebuffi et~al.(2017)Rebuffi, Bilen, and Vedaldi]{rebuffi_arXiv2017_learningmultiplevisualdomains}
Sylvestre-Alvise Rebuffi, Hakan Bilen, and Andrea Vedaldi.
\newblock Learning multiple visual domains with residual adapters.
\newblock In \emph{Advances in Neural Information Processing Systems}, 2017.

\bibitem[Rebuffi et~al.(2018)Rebuffi, Bilen, and Vedaldi]{Rebuffi_Efficient_Multi_Domain_CVPR2018}
Sylvestre-Alvise Rebuffi, Hakan Bilen, and Andrea Vedaldi.
\newblock Efficient parametrization of multi-domain deep neural networks.
\newblock In \emph{Proceedings of the IEEE Conference on Computer Vision and Pattern Recognition (CVPR)}, 2018.

\bibitem[Ridnik et~al.(2021)Ridnik, Ben-Baruch, Noy, and Zelnik-Manor]{ridnik_arXiv2021_imagenet21k_pretraining_masses}
Tal Ridnik, Emanuel Ben-Baruch, Asaf Noy, and Lihi Zelnik-Manor.
\newblock Imagenet-21k pretraining for the masses.
\newblock In \emph{Thirty-fifth Conference on Neural Information Processing Systems Datasets and Benchmarks Track (Round 1)}, 2021.

\bibitem[Rombach et~al.(2022)Rombach, Blattmann, Lorenz, Esser, and Ommer]{Rombach_2022_CVPR_stable_diffusion}
Robin Rombach, Andreas Blattmann, Dominik Lorenz, Patrick Esser, and Bj\"orn Ommer.
\newblock High-resolution image synthesis with latent diffusion models.
\newblock In \emph{Proceedings of the IEEE/CVF Conference on Computer Vision and Pattern Recognition (CVPR)}, pages 10684--10695, 2022.

\bibitem[Selva et~al.(2023)Selva, Johansen, Escalera, Nasrollahi, Moeslund, and Clapés]{Selva_PAMI2023_Video_Trans_Survey}
Javier Selva, Anders~S. Johansen, Sergio Escalera, Kamal Nasrollahi, Thomas~B. Moeslund, and Albert Clapés.
\newblock Video transformers: A survey.
\newblock \emph{IEEE Transactions on Pattern Analysis and Machine Intelligence}, 45\penalty0 (11):\penalty0 12922--12943, 2023.

\bibitem[Soomro et~al.(2012)Soomro, Zamir, and Shah]{Soomro_arXiv2012_UCF101}
Khurram Soomro, Amir~Roshan Zamir, and Mubarak Shah.
\newblock {UCF101:} {A} dataset of 101 human actions classes from videos in the wild.
\newblock \emph{CoRR}, abs/1212.0402, 2012.

\bibitem[Sung et~al.(2022)Sung, Cho, and Bansal]{Sung_VL_Adapter_CVPR2022}
Yi-Lin Sung, Jaemin Cho, and Mohit Bansal.
\newblock {VL-Adapter}: Parameter-efficient transfer learning for vision-and-language tasks.
\newblock In \emph{Proceedings of the IEEE/CVF Conference on Computer Vision and Pattern Recognition (CVPR)}, pages 5227--5237, 2022.

\bibitem[Torralba and Efros(2011)]{Torralba_Unbiased_dataset_bias_CVPR2011}
Antonio Torralba and Alexei~A. Efros.
\newblock Unbiased look at dataset bias.
\newblock In \emph{Proceedings of the IEEE/CVF Conference on Computer Vision and Pattern Recognition (CVPR)}, pages 1521--1528, 2011.
\newblock ISSN: 1063-6919.

\bibitem[Ulhaq et~al.(2022)Ulhaq, Akhtar, Pogrebna, and Mian]{Ulhaq_arXiv2022_Transformers_Action_Recognition_Survey}
Anwaar Ulhaq, Naveed Akhtar, Ganna Pogrebna, and Ajmal Mian.
\newblock Vision transformers for action recognition: {A} survey.
\newblock \emph{CoRR}, abs/2209.05700, 2022.

\bibitem[Vandenhende et~al.(2022)Vandenhende, Georgoulis, Van~Gansbeke, Proesmans, Dai, and Van~Gool]{Vandenhende_Dence_MTL_survey_PAMI2021}
Simon Vandenhende, Stamatios Georgoulis, Wouter Van~Gansbeke, Marc Proesmans, Dengxin Dai, and Luc Van~Gool.
\newblock Multi-task learning for dense prediction tasks: A survey.
\newblock \emph{IEEE Transactions on Pattern Analysis and Machine Intelligence}, 44\penalty0 (7):\penalty0 3614--3633, 2022.

\bibitem[Wang et~al.(2021)Wang, Liu, Wang, and Jing]{Wang_ICCV2021_TesNet}
Jiaqi Wang, Huafeng Liu, Xinyue Wang, and Liping Jing.
\newblock Interpretable image recognition by constructing transparent embedding space.
\newblock In \emph{Proceedings of the IEEE/CVF International Conference on Computer Vision (ICCV)}, pages 895--904, 2021.

\bibitem[Weinzaepfel and Rogez(2021)]{Weinzaepfel_IJCV2021_Mimetics_dataset}
Philippe Weinzaepfel and Gr{\'{e}}gory Rogez.
\newblock Mimetics: Towards understanding human actions out of context.
\newblock \emph{International Journal of Computer Vision}, 129\penalty0 (5):\penalty0 1675--1690, 2021.

\bibitem[Xie et~al.(2024)Xie, Chen, Zhang, Wan, and Li]{Xie_Large_Multimodal_Agents_survey_2024}
Junlin Xie, Zhihong Chen, Ruifei Zhang, Xiang Wan, and Guanbin Li.
\newblock Large multimodal agents: {A} survey.
\newblock \emph{CoRR}, abs/2402.15116, 2024.

\bibitem[Xin et~al.(2024{\natexlab{a}})Xin, Du, Wang, Lin, and Yan]{Xin_VMT_Adapter_AAAI2024}
Yi Xin, Junlong Du, Qiang Wang, Zhiwen Lin, and Ke Yan.
\newblock {VMT}-{Adapter}: {Parameter}-{Efficient} {Transfer} {Learning} for {Multi}-{Task} {Dense} {Scene} {Understanding}.
\newblock In \emph{Proceedings of the {AAAI} {Conference} on {Artificial} {Intelligence}}, pages 16085--16093, 2024{\natexlab{a}}.
\newblock Number: 14.

\bibitem[Xin et~al.(2024{\natexlab{b}})Xin, Luo, Zhou, Du, Liu, Fan, Li, and Du]{Xin_PEFT_survey_2024}
Yi Xin, Siqi Luo, Haodi Zhou, Junlong Du, Xiaohong Liu, Yue Fan, Qing Li, and Yuntao Du.
\newblock Parameter-efficient fine-tuning for pre-trained vision models: {A} survey.
\newblock \emph{CoRR}, abs/2402.02242, 2024{\natexlab{b}}.

\bibitem[Xing et~al.(2024)Xing, Liu, Wang, Sun, Chen, Gu, and Wang]{Xing_PEFT_survey_2024}
Jialu Xing, Jianping Liu, Jian Wang, Lulu Sun, Xi Chen, Xunxun Gu, and Yingfei Wang.
\newblock A survey of efficient fine-tuning methods for {Vision}-{Language} {Models} — {Prompt} and {Adapter}.
\newblock \emph{Computers \& Graphics}, 119:\penalty0 103885, 2024.

\bibitem[Xu et~al.(2023)Xu, Zhu, and Clifton]{Xu_Multimodal_survey_PAMI2023}
Peng Xu, Xiatian Zhu, and David~A. Clifton.
\newblock Multimodal {Learning} {With} {Transformers}: {A} {Survey}.
\newblock \emph{IEEE Transactions on Pattern Analysis and Machine Intelligence}, 45\penalty0 (10):\penalty0 12113--12132, 2023.

\bibitem[Yipeng(2022)]{tensor_book2021}
Liu Yipeng, editor.
\newblock \emph{Tensors for Data Processing: Theory, Methods, and Applications}.
\newblock Academic Press, 1st edition, 2022.

\bibitem[Zhang et~al.(2024)Zhang, Han, Liu, Zhou, Lu, Qiao, Li, and Gao]{Zhang_LLaMA_Adapter_ICLR2024}
Renrui Zhang, Jiaming Han, Chris Liu, Aojun Zhou, Pan Lu, Yu Qiao, Hongsheng Li, and Peng Gao.
\newblock {LL}a{MA}-adapter: Efficient fine-tuning of large language models with zero-initialized attention.
\newblock In \emph{The Twelfth International Conference on Learning Representations}, 2024.

\bibitem[Zhang and Yang(2022)]{Zhang_MTL_Survey_2021}
Yu Zhang and Qiang Yang.
\newblock A survey on multi-task learning.
\newblock \emph{IEEE Transactions on Knowledge and Data Engineering}, 34\penalty0 (12):\penalty0 5586--5609, 2022.

\bibitem[Zhou et~al.(2022)Zhou, Koltun, and Kr\"ahenb\"uhl]{Zhou_Simple_MultiDataset_CVPR2022}
Xingyi Zhou, Vladlen Koltun, and Philipp Kr\"ahenb\"uhl.
\newblock Simple multi-dataset detection.
\newblock In \emph{Proceedings of the IEEE/CVF Conference on Computer Vision and Pattern Recognition (CVPR)}, pages 7571--7580, 2022.

\end{thebibliography}
}


\end{document}